\documentclass[10pt,twocolumn,letterpaper]{article}

\usepackage{wacv}
\usepackage{times}
\usepackage{epsfig}
\usepackage{graphicx}
\usepackage{amsmath}
\usepackage{amssymb}
\usepackage{color}
\usepackage{comment}
\usepackage{multirow}
\usepackage{rotating}
\usepackage{flushend} 
\usepackage[resetlabels]{multibib}

\definecolor{purple}{rgb}{.75,0,1}

\newif\ifcomments
\newcounter{cmtct}
\wacvfinalcopy 

\ifwacvfinal\fi
\newcites{adx}{Appendix-Literature}
\begin{document}

\title{Deep Learning Prototype Domains for Person Re-Identification}

\author{
Arne Schumann\\
Fraunhofer IOSB, Karlsruhe, Germany\\
{\tt\small arne.schumann@iosb.fraunhofer.de}
\and
Shaogang Gong \\
Queen Mary University of London, UK\\
{\tt\small s.gong@qmul.ac.uk}
\and
Tobias Schuchert\\
Fraunhofer IOSB, Karlsruhe, Germany\\
{\tt\small tobias.schuchert@iosb.fraunhofer.de}
}

\maketitle
\ifwacvfinal\thispagestyle{empty}\fi

\begin{abstract}
    Person re-identification (re-id) is the task of matching multiple
    occurrences of the same person from different cameras, poses,
    lighting conditions, and a multitude of other factors which alter
    the visual appearance.  Typically, this is achieved by learning
    either optimal features or matching metrics which are adapted to
    specific pairs of camera views dictated by the pairwise labelled
    training datasets.  In this work, we formulate a deep learning
    based novel approach to automatic {\em prototype-domain} discovery
    for domain perceptive (adaptive) person re-id (rather than camera
    pair specific learning) for any camera views scalable to new
    unseen scenes without training data. We learn a separate
    re-id model for each of the discovered prototype-domains and
    during model deployment, use the person probe image to select
    {\em automatically} the model of the closest prototype-domain. Our
    approach requires neither supervised nor unsupervised domain
    adaptation learning, i.e. no data available from the target
    domains.  We evaluate extensively our model under realistic
    re-id conditions using automatically detected bounding boxes
    with low-resolution and partial occlusion. We show that our
    approach outperforms most of the state-of-the-art supervised and
    unsupervised methods on the latest CUHK-SYSU and PRW benchmarks.  
\end{abstract}

\section{Introduction}
\label{sec:introduction}

The task of re-identifying the same person across different cameras has
attracted much interest in recent years. Person re-identification is at its
core a cross-domain recognition problem. Datasets are usually recorded in a
camera network setting with a fixed set of cameras and viewing angles.
Consequently, most approaches interpret each camera as a separate visual domain
and focus on developing features or metrics that can robustly recognize a
person within such {\em camera-view-perspective} domains. In this work, we
consider other {\em camera-view-independent} factors, such as pose,
illumination,  
occlusions, and background influence the visual appearance of a
person, and we wish to explore them as visual domains in constructing
{\em camera-view independent re-id} models for better scalability to
unknown camera views.

\begin{figure}[t]
    \begin{center}
        \includegraphics[width=0.99\linewidth]{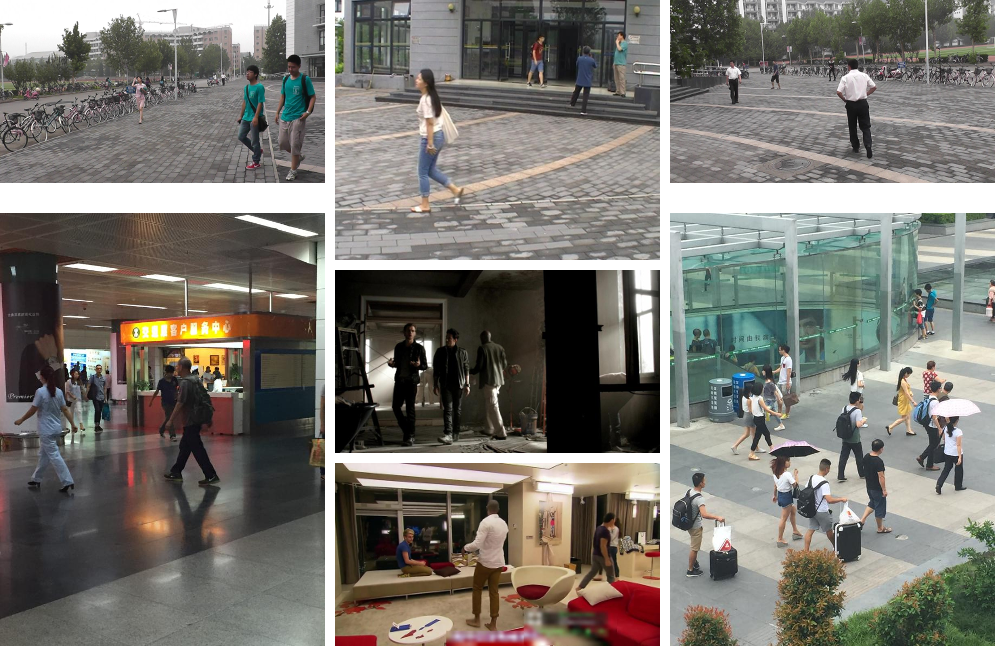}
    \end{center}
    \caption{We use the latest PRW \cite{zheng2016person} (top row) and CUHK-SYSU
             \cite{xiao2016end} (bottom row) datasets as target (test)
             domains for evaluation, unavailable to model
             learning. Both datasets provide many camera 
             views and unsegmented video frames which requires
             auto-person-detection for a more realistic person re-id
             evaluation. }
    \label{fig:intro}
\end{figure}

In this work, we propose a two-stage approach to automatically discover visual
domains in large amounts of diverse data and use them to learn feature
embeddings for person re-identification (see Figure \ref{fig:intro}).
In the first stage, we pool data from a large amount of re-identification
datasets, independent from the test domains, to capture a
large degree of visual variation in the training data. We then explore
clustering based on feature learning in convolutional neural
networks (CNNs) to automatically discover dominant (prototype) visual domains.
In the second stage, we again apply CNNs to learn feature embeddings
in each of the prototype domains in order to support domain perceptive
(sensitive) person re-id during testing with automatic
domain selection. We learn one embedding 
per domain. This allows the model to learn specific details about
each individual prototype domain while ignoring others. For example,
an embedding learned for a domain which predominantly contains people
of dark-dress does not need to encode information relevant to
distinguishing a person dressed in light blue colors from a person
dressed in white clothes. By doing so, the domain perceptive embedding
focuses on learning subtle discriminative characteristics among
similar visual appearances.
On testing, a probe image is first matched to its most likely domain. Then, the
feature embedding learned on that domain is used to perform re-identification.
%
Note, this approach is purely {\em inductive}. It does not require any training
data (labelled or unlabelled) from the target (test) domains,
and the model is designed to scale to any new target domains.
%
Our approach is particularly well suited to scenarios in which no fixed set of
camera views is available (i.e. no fixed domain borders are specified). We thus
evaluate it on the latest CUHK-SYSU and PRW datasets, which contain
images from diverse sources of mobile cameras, movies and fixed view
cameras, with multitude of view angles, backgrounds, resolutions 
and poses. Our approach yields the state-of-the-art accuracy on
CUHK-SYSU and is competitive on PRW. This is {\em without} using
target domain data in our model training whilst {\em all} other methods
compared in the evaluation exploit target domain data in their model learning.

Our contributions are: 
\textbf{(1)} We formulate a novel approach to automatic discovery of
prototype-domains, characterising person visual appearance with domain
perceptive awareness.
\textbf{(2)} We develop a deep learning model for domain perceptive (DLDP)
selection and re-id matching in a single automatic process without any
supervised nor unsupervised domain transfer learning.
\textbf{(3)} We show the significant advantage of our model by outperforming
the state-of-the-art on the CUHK-SYSU benchmark \cite{xiao2016end} with up
to 5.6\% at Rank-1 re-id, and being competitive on the PRW
benchmark \cite{zheng2016person} of 45.4\% Rank-1 re-id compared to the
47.7\% state-of-the-art, notwithstanding that the latter benefited from
model learning on target domain data.

\section{Related Work}
\label{sec:relwork}

Most re-id approaches can be grouped into two categories: feature based
approaches and metric based approaches. The former type aims to develop a
robust feature representation. The latter
approach focuses on optimizing a distance metric that, given any
feature, yields small distances for matching person images and large
distances of images of different people.
In recent years, deep learning methods have gained significant
advantages on image classification, and have been applied to person
re-id. 
Many deep learning approaches focus on feature learning. Yi \etal 
\cite{yi2014deep} split person images into three regions and learn 
separate feature maps which are combined into a final feature through a fully
connected layer. Cosine distance is used to perform re-id. Ding \etal
\cite{ding2015deep} apply a triplet loss in order to train a feature whose
Euclidean distance of a matching image pair is smaller than that of a pair of
images of different people. Xiao \etal \cite{xiao2016learning} propose
to use dataset specific dropout to learn features over multiple smaller
datasets simultaneously. Cheng \etal \cite{cheng2016person} propose an
improved triplet loss function which emphasises small distances for similar
image pairs. We \etal \cite{wu2016enhanced} combine hand-crafted features
with CNN features for re-id metric learning.
Other deep learning approaches focus on studying network layers specifically
designed for person re-id. Li \etal \cite{li2014deepreid} describe a filter
pairing network to model translation, occlusion and
background clutter in its architecture. Ahmend \etal \cite{ahmed2015improved}
introduce a neighborhood matching layer for improving robustness to
translation and pose change. This layer is also applied by Wu \etal
\cite{wu2016personnet} to train an end-to-end re-id net which directly outputs
a \mbox{(dis-)} similarity decision without relying on a separate distance 
function.
Xiao \etal \cite{xiao2016end} propose an approach which combines person
detection and re-id into a single CNN for simultaneous person
detection and computing re-id feature for each detection.

A few studies have addressed cross-domain re-id by using target
domain data for supervised
\cite{layne2013domain,ma2014person,wang2015cross} or unsupervised
\cite{ma2015cross,peng2016unsupervised} domain adaptation. 
Others have evaluated their models on datasets without any adaptation 
to the target domain
\cite{hu2014cross,mclaughlin2015data,yi2014deep}. To our knowledge, 
the proposed model in this work, for the first time, 
does not rely on domain adaptation using target domain
data whilst learning domain perceptive re-id for unknown target
domains.

\section{Methodology}
\label{sec:methods}

\begin{figure*}[t]
    \begin{center}
        \includegraphics[width=\textwidth]{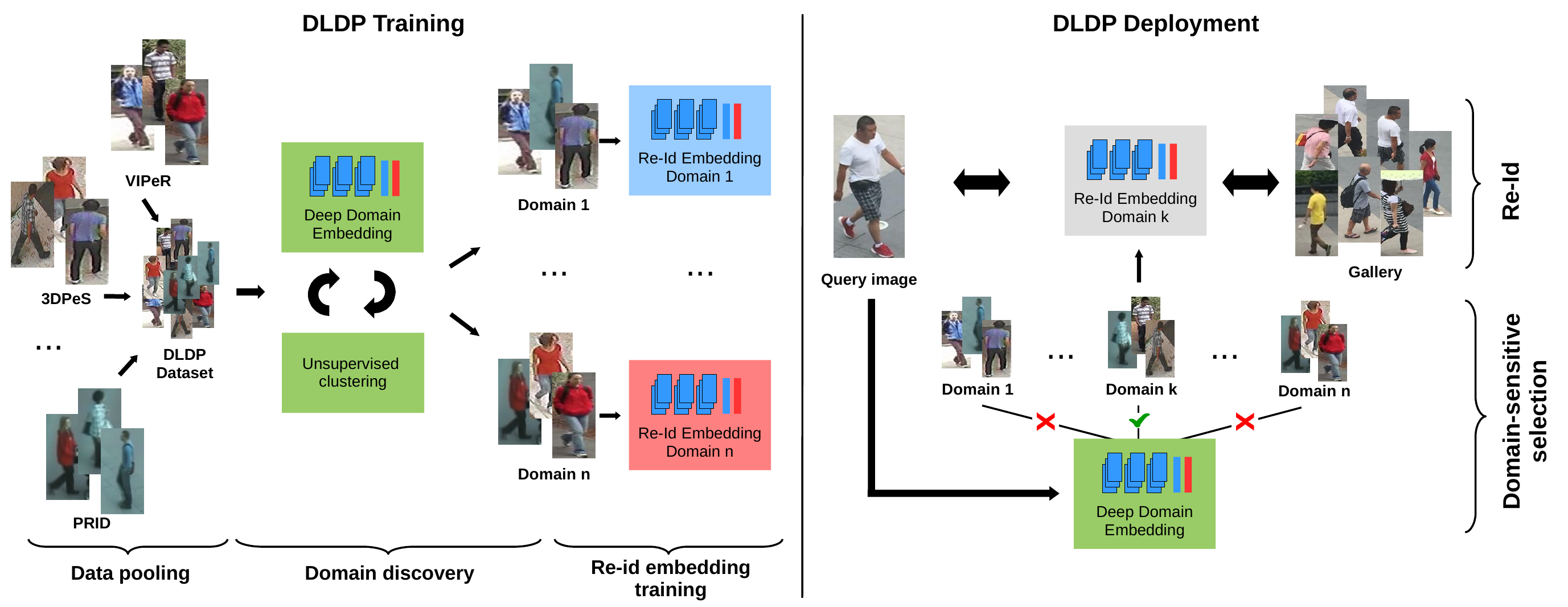}
    \end{center}
    \caption{Overview of DLDP. During training we discover domains in a large pooled dataset. For 
             each domain a domain-specific re-id model is trained. At deployment, the query image is 
             used to identify the closest matching domain and use the corresponding domain-specific 
             model to rank the gallery.}
    \label{fig:overview}
\end{figure*}

The central objective of our approach is to learn a domain adaptive re-id model
(domain perceptive) which is scalable to new and unseen data without requiring
any manual labelling for model training on the new target domains. We propose a
two-stage approach to achieve this: (1) In the first stage, characteristic and
dominant (prototype) domains are automatically discovered in large amounts of
diverse data (Section 3.1.2); (2) In the second stage, this information is used
to train a number of domain specific embeddings by deep learning for person
re-id (Section \ref{sec:domain_embeddings}). An overview of our approach is 
given in Figure \ref{fig:overview}.

\subsection{Automatic Domain Discovery}
\label{sec:discovery}


\subsubsection{Divergent Data Sampling}

A key requirement for a meaningful domain discovery is {\em divergent data
sampling} which aims to provide a large range of realistic visual variation. In
order to achieve such a high degree of variation, we pool a number of publicly
available person re-identification datasets into a new, large dataset for domain
discovery, called DLDP domain discovery dataset\footnote{The DLDP dataset will
be made publically available.}. We combine 10 datasets which together contain
images of 4,786 different persons with a total of 41,380 bounding boxes.
Of these bounding boxes 27,283 are manually annotated and 14,097 are obtained
by a person detector. Table \ref{tab:data_pool} shows the sources used to 
construct the DLDP domain discovery dataset. We resize all bounding boxes to a 
uniform size of $160\times80$ pixels. This DLDP divergent data sampling allows 
us to discover domains which cover a large and diverse spectrum of possible 
variation in person visual appearance. We show in our experiments its 
suitability for generalisation to person re-id in new/unseen camera domains.

\begin{table}
\begin{center}
\resizebox{.48\textwidth}{!}{
\begin{tabular}{|l|c|c|c|c|}
\hline
                                        & Persons   & Cameras & M-BBoxes & A-BBoxes \\
\hline\hline
HDA \cite{figueira2014hda+}             &    85     & 13      &    850   & -        \\ 
GRID \cite{loy2010time}                 &   250     &  8      &    500   & -        \\ 
3DPeS \cite{baltieri20113dpes}          &   200     &  8      &  1,012   & -        \\ 
CAVIAR4REID \cite{Cheng_BMVC11}         &    72     &  2      &  1,221   & -        \\ 
i-LIDS \cite{ilids}                     &   119     &  2      &    476   & -        \\ 
PRID \cite{hirzer2011person}            &   200     &  2      &    400   & -        \\ 
VIPeR \cite{gray2008viewpoint}          &   632     &  2      &  1,264   & -        \\ 
SARC3D \cite{baltieri2011sarc3d}        &    50     &  1      &    200   & -        \\ 
CUHK2 \cite{li2013locally}              & 1,816     & 10      &  7,264   & -        \\ 
CUHK3 \cite{li2014deepreid}             & 1,360     &  6      & 14,096   & 14,097   \\
\hline\hline
Total       & 4,786     & 54      & 27,283   & 14,097   \\
\hline
\end{tabular}
}
\vspace{.1cm}
\caption{Ten sources for the DLDP re-id domain discovery dataset. The data consists of manually 
labelled bounding boxes (M-BBoxes) as well as person detections (A-BBoxes).}
\label{tab:data_pool}
\end{center}
\end{table}

\subsubsection{Prototype-Domain Discovery}
\label{sec:pd_discovery}

We wish to explore deep learning based clustering to discover dominant
(prototype) visual domains from the multi-source pooled DLDP dataset. In
particular, we exploit the concept of unsupervised deep embedding space
learning proposed in \cite{xie2015unsupervised}, but importantly,
adopted to utilise the available person id labels from the re-id
datasets. Our {\em supervised} deep 
learning clustering model alternates between (1) training a CNN to learn a
feature embedding from the re-id image datasets and (2) applying conventional
k-means clustering in the embedding space to find clusters.
To initialize the weights of our feature embedding CNN, we train the model
using the person ID labels available in the data. Our model architecture is
given in Table~\ref{tab:architecture} (Section \ref{sec:domain_embeddings} 
for more details). We set the last, fully connected layer of the network to 
4,786 dimensions and train using person ID labels in a one-hot encoding and a 
softmax loss for person ID classification.
The multi-source dataset ensures that the influence of any particular dataset's 
bias on the initial feature embedding is reduced. Moreover, we apply data
augmentation by cropping and flipping the images, and also ensure unbiased
sampling by selecting images from different data sources in DLDP with equal
frequency. Image cropping is performed by resizing an image to $30\times10$ 
pixels larger than the net requires and randomly cropping it down to the correct 
size. Through data augmentation, the hypothetical data pool size is increased by 
a factor of 600. We ensure unbiased sampling by selecting images from different 
data sources in DLDP with equal frequency. This results in more data 
augmentation on the smaller data sources (among the ten sources in the DLDP 
dataset) so to prevent the CNN from overfitting to the larger data sources. 
For the domain discovery part (k-means clustering) the person ID softmax loss 
layer is replaced by a softmax loss which corresponds to the number of clusters, 
set to eight in our current model\footnote{We choose 8 clusters as
  a tradeoff between the number of domains available to our
  model and the computational effort involved in training our
  domain-specific embeddings. Future work can further optimise it.}.
Thus, after a supervised initialization, the domain discovery continues in an 
unsupervised manner.

The initialization from a person re-identification net is crucial to the
success of our prototype domain discovery. The re-identification training
ensures that the initial model does {\em not} react strongly to the dataset
biases present in our feature pool. {\em This prevents the clustering from
simply discovering trivial dataset boundaries as prototype domain boundaries} 
and instead, lets the model focuses more on the content of each person bounding 
box.

\subsubsection{Training Strategy}

\begin{figure}[t]
\begin{center}
\includegraphics[width=0.99\linewidth]{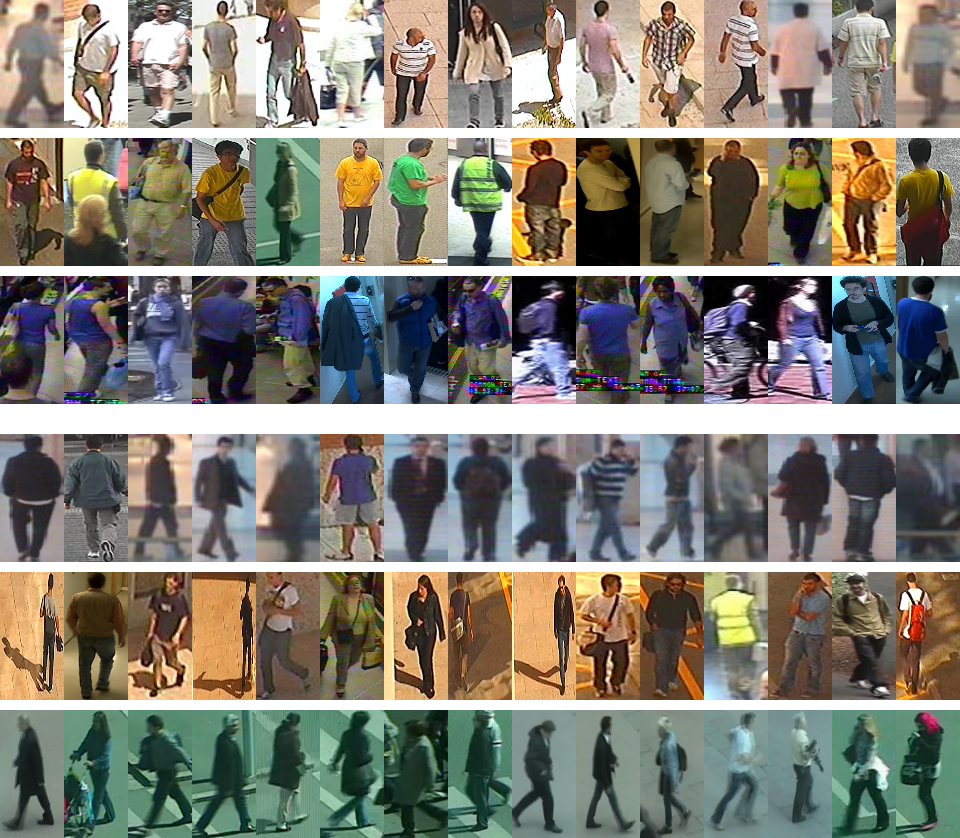}
\end{center}
\caption{Example domains discovered by our approach using the proposed 
         initialization with a re-id net (top 3 rows, supervised 
         initialization) and initialization by weights learned through 
         autoencoding (AE) (bottom 3 rows, unsupervised initialization).
         The re-id initialization leads to more semantically meaningful
         domain (\eg light-colored, yellow and blue clothing). The AE 
         initialization is strongly influenced by dataset bias and learns
     domains corresponding to datasets (\eg CAVIAR4REID, 3DPeS, PRID).}
    \label{fig:clustering}
\end{figure}

For training of our deep clustering model we use a low initial learning rate of
0.001. This ensures that the cluster embedding does not deviate too quickly
from its re-id label constrained initialization. Given the initial embedding,
we perform 25 runs of k-means clustering in the embedding space and select the 
best result for the next refinement of the embedding (\ie step 2 in Section 
\ref{sec:pd_discovery}). This ensures stability of the iterative training 
process. The refinement (fine-tuning) of the embedding CNN is then performed
for a further 10,000 training iterations (\ie step 1 in Section
\ref{sec:pd_discovery}). We divide the learning rate of the embedding by 10 
every two iterations of the discovery process. This iterative process is
repeated until less than 1\% of images change their cluster assignments.
Some examples of learned prototype domains (\ie clusters in the embedding 
feature space) with their corresponding images are shown in Figure
\ref{fig:clustering}.

\subsection{Deep Learning Domain Perceptive Re-Id Model}
\label{sec:domainmodel}

The second stage of our DLDP re-id model consists of learning a
domain-sensitive re-id model for each prototype domain. That is, we train one
feature embedding {\em with all person ID labels} for each of the discovered
clusters in the feature embedding space resulted from the first stage. To that
end, we start by training a common generic baseline re-id model on all
available data without considering the domains. The individual domain models
are then trained by fine-tuning this baseline model. The same baseline model is
also used as initialization for the domain discovery approach described in
Section \ref{sec:pd_discovery}.

\subsubsection{Baseline Generic Model}
\label{sec:baseline}

As a baseline approach we train a model of the architecture given in
Table~\ref{tab:architecture} (Section \ref{sec:domain_embeddings} for more
details) on all available training data to learn a generic feature embedding
without domain specific adaptation. We train the baseline model for 60,000
iterations. The initial learning rate is set to 0.1 and divided by 10 after
every 20,000 iterations. We use the output of the 512 dimensional layer (fc
feat in Table~\ref{tab:architecture}) just before the loss as our feature
embedding for person re-id. The resulting features are compared using cosine
distance.

\newcommand{\by}{$\times$}
\newcommand{\specialcell}[2][c]{\begin{tabular}[#1]{@{}c@{}}#2\end{tabular}}
\begin{table}[ht]
\vspace{1cm}
\begin{center}       
\begin{tabular}{|l|c|c|c|c|c|c|c|}
\hline
  name            
& \specialcell{patch size,\\stride} 
& output dim               
& \# filters \\
\hline
\hline
input           &                   &   3 \by{} 160 \by{} 64 &      \\
\hline
conv 1-4        & 3 \by{} 3, 1      &  32 \by{} 160 \by{} 64 &      \\
\hline
pool            & 2 \by{} 2, 2      &  32 \by{}  80 \by{} 32 &      \\
\hline
inception 1a    &                   & 256 \by{}  80 \by{} 32 &  64  \\
\hline
inception 1b    & stride 2          & 384 \by{}  40 \by{} 16 &  64  \\
\hline
inception 2a    &                   & 512 \by{}  40 \by{} 16 & 128  \\
\hline
inception 2b    & stride 2          & 768 \by{}  20 \by{} 8  & 128  \\
\hline
inception 3a    &                   & 1024\by{}  20 \by{} 8  & 128  \\
\hline
inception 3b    & stride 2          & 1536\by{}  10 \by{} 4  & 128  \\
\hline
fc feat         &                   & 512                    &      \\
\hline
fc loss         &                   & \#person ids           &      \\
\hline
\end{tabular}
\end{center}
\caption{DLDP model architecture for prototype domain discovery.}
\label{tab:architecture}
\end{table}

\subsubsection{Domain Embeddings}
\label{sec:domain_embeddings}

In order to learn feature embedding focused on each of the domains we need to
first create suitable domain-specific training data. For any person ID in a
given domain we thus select all of that person's images and add them to the
training data for the domain. This data sampling method allows the domain
models to specialize and focus particularly on the visual cues relevant to
persons from their domain while not having to also learn how to distinguish
persons from different domains.

The architecture (Table~\ref{tab:architecture}) we use to learn the
domain-specific feature embedding is motivated by a number of recent studies.
It consists of an inital set of four convolutional layers with filter sizes of
$3\times3$. This configuration of multiple layers with small filter sizes was
shown to perform well for image classification in the VGG nets
\cite{simonyan2014very}. We further adopt insights from
\cite{szegedy2015going} and \cite{szegedy2015rethinking} to add multiple (four)
inception layers to our network. We modify the original inception architecture
by replacing the $5\times5$ layer with two $3\times3$ layers, reducing the grid
size and expanding filter banks as suggested in \cite{szegedy2015rethinking}.
We apply batch normalization \cite{ioffe2015batch} after each layer and use a
softmax loss based on the person ID labels for training. Our feature embeddings
are of size 512.

\subsubsection{Training Strategy}

We begin training by disregarding the identified domain borders and combining
all available person IDs into one softmax layer. We train this net for an
initial 60,000 iterations with a learning rate on 0.1 which is divided
by 10 every 20,000 iterations. After this, we continue to train individually
for each domain relying only the corresponding data pool. The dimension of the
softmax layers is adapted accordingly. For each domain we continue training for
30,000 iterations at an inital learning rate of 0.001. Our input images are
resized to a size of $210\times70$. Data augmentation is then performed by
randomly flipping images and randomly cropping them to a final input size of
$180\times60$. Similar to \cite{wu2016enhanced} we apply hard negative mining
by selecting misclassified training images and fine-tuning each net for a
further 10,000 iterations at a reduced learning rate of 0.00001.

\subsubsection{Automatic Domain Selection}
\label{sec:selection}

After model training and during model deployment, a probe person image is first
matched to its most likely domain by the deep clustering model (Section
\ref{sec:discovery}). The corresponding domain embedding (domain specific re-id
model) is then used to rank the gallery images by computing the corresponding
512 dimensional embedding and using cosine distance for matching the probe
image. Note, the camera view and the target domain of the probe image is new,
\ie unseen and independent from any of the multi-sources used to construct the
DLDP dataset.

\section{Experiments}

\newcommand{\f}[1]{\textbf{#1}}          
\newcommand{\s}[1]{\textit{\textbf{#1}}} 

\noindent {\bf Datasets}: 
We evaluate our model on two publicly available large re-id datasets: CUHK-SYSU
\cite{xiao2016end} and PRW \cite{zheng2016person}, both of which are
independent/unseen from the ten multi-source data pool used to construct our
DLDP domain discovery training dataset.
Both datasets contain a large number of viewing angles. CUHK-SYSU consists of
pedestrian images collected by handheld cameras as well as scenes from movies
and the PRW dataset was collected with six cameras on a campus environment. The
datasets contain 8432 and 932 person ids and 99,809 and 34,304 bounding boxes,
respectively. Both datasets provide full images to enable automatically
detected person bounding boxes to be evaluated in person re-id, subject to
occlusion, bbox misalignment, and large changes in
resolution/low-resolution. Some example images of both datasets are depicted in
Figure \ref{fig:intro}.
These characteristics of the two datasets allow us to investigate the
generalization capability of our approach, its ability to handle large amounts
of varying views and to evaluate its performance against automatically detected
person bboxes for more realistic evaluation.

\noindent {\bf Evaluation protocol}: 
A central objective of our approach is {\em not} to require any training data
on the target domain for the re-id task. To that end, in the experiments we
only used the test part of both datasets. The CUHK-SYSU dataset contains a
fixed set of 2,900 query persons and gallery sets of multiple sizes (at most
6,978 images). The PRW dataset contains a fixed query set of 2,057 bounding
boxes and a gallery size of 6,112 test images. Note that in both datasets each
gallery image contains multiple persons and an automatic person detector may
generate additional false posistive bounding boxes. We follow the exact
evaluation protocols specified in \cite{xiao2016end} and \cite{zheng2016person}
respectively, and used the provided evaluation code where applicable. Also
note, both datasets contain many persons without id in the galleries, \ie the
re-id tasks in these datasets are potentially {\em open-set} given the unknown
distractors in the target population. To give a direct comparison to the
reported results in \cite{xiao2016end} and \cite{zheng2016person}, we also
adopt mean Averaged Precision (mAP) and Rank-1 accuracy as evaluation metrics.

\begin{table}[ht!]
\begin{center}
\begin{tabular}{|c|l|c|c|}
\cline{3-4}
\multicolumn{2}{c|}{}
                                         & mAP       & Rank-1   \\
\cline{3-4}
\noalign{\vskip .08cm}
\hline
\multirow{6}{*}{\begin{turn}{90}GT\end{turn}}  
&Euclidean \cite{xiao2016end}            & 41.1      & 45.9     \\
&KISSME \cite{kostinger2012large}        & 56.2      & 61.9     \\
&BoW \cite{zheng2015scalable}            & 62.5      & 67.2     \\
&IDNet \cite{xiao2016end}                & 66.5      & 71.1     \\
&Baseline Model                          & 68.4      & 70.3     \\
&DLDP                                    & \f{74.0}  & \f{76.7} \\
\hline\hline
\multirow{4}{*}{\begin{turn}{90}\hspace{-.2cm}Detections\end{turn}}
&Person Search \cite{xiao2016end}        & 55.7      & 62.7     \\
&Person Search rerun                     & 55.79     & 62.17    \\
&DLDP (SSD VOC300)                       & 49.53     & 57.48    \\
&DLDP (SSD VOC500)                       & \f{57.76} & \f{64.59}\\
\hline
\end{tabular}
\vspace{.1cm}
\caption{DLDP re-id performance comparison against both supervised
         (KISSME, IDNet, Person Search) and unsupervised (Euclidean, BoW) 
         methods on the CUHK-SYSU dataset.} 
\label{tab:eval_sota_cuhk}
\end{center}
\end{table}

\begin{table}[ht!]
\begin{center}
\begin{tabular}{|c|l|c|c|}
\cline{3-4}
\multicolumn{2}{c|}{}
                                         & mAP       & Rank-1    \\
\cline{3-4}
\noalign{\vskip .08cm}
\hline
\multirow{5}{*}{\begin{turn}{90}\hspace{-.1cm}DPM Inria\end{turn}}  
&IDE \cite{zheng2016person}              &    13.7   &    38.0   \\
&IDE$_{det}$ \cite{zheng2016person}      & \f{18.8}  & \f{47.7}  \\
&BoW + XQDA \cite{zheng2016person}       &    12.1   &    36.2   \\
&Baseline Model                          &    12.9   &    36.5   \\
&DLDP                                    &    15.9   &    45.4   \\
\hline
\hline
\multirow{3}{*}{\begin{turn}{90}\hspace{-.1cm}SSD\end{turn}}  
&BoW + XQDA (SSD VOC300)                 &     6.8   &    26.6   \\
&DLDP (SSD VOC300)                       &    10.1   &    35.3   \\
&DLDP (SSD VOC500)                       & \f{11.8}  & \f{37.8}  \\
\hline

\end{tabular}
\vspace{.1cm}
\caption{DLDP re-id performance on the PRW dataset in comparison to
         state-of-the-art. All results are obtained by considering 5 bounding
         boxes per image. Note that all approaches except ours were trained
         (supervised) on the PRW dataset.}
\label{tab:eval_sota_prw}
\end{center}
\end{table}

\noindent {\bf Comparison with the state-of-the-art}: 
To demonstrate the effectiveness of our approach, we compared our model directly
to the state-of-the-art reported in  \cite{xiao2016end} and
\cite{zheng2016person}, using both manually labelled person bounding boxes
(ground truth) and automatically detected bounding boxes.
Results on the CUHK-SYSU dataset for gallery sizes of 100 images are given in
Table \ref{tab:eval_sota_cuhk}. Our baseline generic re-id model (Section 
\ref{sec:baseline}) given manually labelled person bounding boxes (ground truth)
as input outperforms not only \cite{xiao2016end} using conventional image
features but also the deep IDNet model which has the advantage of being trained
on the CUHK-SYSU dataset itself at Rank-1 by 1.9\%. The reason is likely a
combination of our deeper 10 layer network architecture, the use of inception
layers and batch normalization.
For our domain adaptive model given manually labelled person bounding boxes,
our model outperforms \cite{xiao2016end} by 7.5\% and 5.6\% in mAP and Rank-1
respectively, a further improvement of 6\% in both mAP and Rank-1 over our
generic baseline model. This suggests that the DLDP model learning for
prototype-domain adaptive re-id is more effective than the ``blind'' generic
model.

For automatic detection generated person bounding boxes, we adopt the
SSD VOC500 person detector \cite{liu2015ssd}. For re-id given these
automatic detections, our prototype-domain adaptive model outperforms the
state-of-the-art person search deep model \cite{xiao2016end} by 2.06\% and
1.89\% on mAP and Rank-1 respectively, despite a very critical
difference that the person search 
deep model \cite{xiao2016end} was trained jointly for person detection
and re-identification {\em using part of the CUHK-SYSU dataset}, \ie
both their detector and their re-id matching model were trained and
fine-tuned on the target domain. In contrast, our DLDP model did not
benefit from training detectors in the target domain, nor fine-tuning
re-id model on the target domain data.

For the evaluation on the PRW benchmark, we compared DLDP to 
a baseline using BoW features and XQDA metric learing \cite{liao2015person} and 
two deep feature embeddings IDE and IDE$_{det}$ from \cite{zheng2016person} 
which are based on the AlexNet \cite{alexnet} architecture, trained on ImageNet 
and fine-tuned for re-id on PRW.
For person detection, we used both the DPM person detector 
\cite{felzenszwalb2010object} trained on the INRIA dataset \cite{dalal2005histograms}
provided by \cite{zheng2016person} and the SSD detectors for a fair
comparison. Our results are shown in Table \ref{tab:eval_sota_prw}. All reported 
results were obtained by considering five bounding boxes per gallery image which 
is the value at which the methods reported in \cite{zheng2016person} perform best.
It is evident that the SSD detectors decrease re-id performance for all models
as the SSD detectors seem to perform poorly on the PRW dataset. Regardless, our
model outperforms both the BOW+XQDA baseline and the deep IDE feature
embedding reported in \cite{zheng2016person} when identical DPM person
detector was used, by 2.2\% and 7.4\% in mAP and Rank-1, respectively (for 
the more accurate DPM detections).
The improved deep IDE$_{det}$ embedding of \cite{zheng2016person} is trained by
fine-tuning the AlexNet first for person/background classification followed by
further fine-tuning for re-id. It outperforms DLDP by 2.9\% and 2.3\% in
mAP and Rank-1 accuracy. However, our performance remains competitive
and has its unique advantage over IDE$_{det}$ embedding. This is
because that IDE$_{det}$ was not only {\em trained directly on PRW} but also
{\em specifically adapted on the PRW} for sensitivity to false
positive person detections. DLDP benefited from none of that.

\begin{table*}[ht!]
\begin{center}
\begin{tabular}{|l|c|c|c|c|c|c|c|c|}
\hline
                                 &       & 50     & 100    & 500    & 1000   & 2000   & 4000   & all (6978)  \\
\hline\hline    
Person Search \cite{xiao2016end} & \multirow{2}{*}{mAP}   
                                         &   58.72&   55.79&   47.38&   43.41&   39.14&   35.70&   32.69     \\
DLDP (SSD VOC500)                &       &\f{60.8}&\f{57.7}&\f{49.2}&\f{45.2}&\f{41.4}&\f{38.1}&\f{35.2}     \\
\hline
Person Search \cite{xiao2016end} & \multirow{2}{*}{Rank-1} 
                                         &   64.83&   62.17&   53.76&   49.86&   45.21&   41.86&   38.69    \\
DLDP (SSD VOC500)                &       &\f{67.6}&\f{64.6}&\f{57.0}&\f{52.9}&\f{49.2}&\f{46.1}&\f{43.1}    \\
\hline
\end{tabular}
\vspace{.1cm}
\caption{Comparison of DLDP to \cite{xiao2016end} for different
  gallery sizes on CUHK-SYSU. 
Results of \cite{xiao2016end} were obtained using the provided code.}
\label{tab:eval_galsize}
\end{center}
\end{table*}

\begin{table*}[ht!]
\begin{center}
\resizebox{.98\textwidth}{!}{
\begin{tabular}{|l|c|c|c|c|c|c|c|c|c|c|}
\hline
&\multicolumn{2}{c|}{Deep+Kissme \cite{xiao2016end}}&\multicolumn{2}{c|}{ACF+BOW \cite{xiao2016end}}&
\multicolumn{2}{c|}{Person Search \cite{xiao2016end}}&
\multicolumn{2}{c|}{Person Search \cite{xiao2016end} rerun}&\multicolumn{2}{c|}{DLDP (SSD VOC500)} \\
\hline
            & mAP   & top-1 & mAP    & top-1 & mAP  & top-1  & mAP   & top-1 & mAP   & top-1 \\
\hline\hline    
Whole       & 39.1  & 44.9  & 42.4   & 48.4  & 55.7 & 62.7   & 55.79 & 62.17 &\f{57.7}&\f{64.6}\\
Occlusion   & 18.2  & 17.7  & 29.1   & 32.1  &\f{39.7}&\f{43.3}& 34.97 & 37.43 & 38.9  & 39.0  \\
LowRes      & 43.8  & 42.8  & 44.9   & 53.8  & 35.7 & 42.1   & 35.21 & 40.00 &\f{41.9} &\f{49.0}\\
\hline
\end{tabular}
}
\vspace{.1cm}
\caption{Comparisons on the CUHK-SYSU occlusion and low resolution tests.}
\label{tab:eval_occres}
\end{center}
\end{table*}

In summary, our DLDP model (given the output from comparable/identical person
detectors) outperforms most of the state-of-the-art person re-identification
methods on both the CUHK-SYSU and the PRW benchmark datasets. It is even more
significant that our results are obtained without any labelled data training on
the target test domains whilst all other methods require training data from the
target domains. Qualitative examples on both datasets are shown in
Figure \ref{fig:qualitative}, including two failure cases in the last row.
Note that the incorrect results for all queries have a color composition or 
clothing configuration that is reasonably similar to the query image. In 
particular, DLPD understandably ranks near-identically looking people 
(PRW, row 2) very high. In the failure case on PRW our model appears to focus 
on the structural pattern created by the bikes in combination with white-dressed 
persons. 

\noindent {\bf Effects from gallery size increase}:
The CUHK-SYSU dataset offers multiple gallery sets of varying sizes. This
allows us to evaluate the influence of larger numbers of detractors on
re-identification accuracy in a more realistic open-set setting. Table
\ref{tab:eval_galsize} shows results of our DLDP model in comparison to those
achieved by the end-to-end person search deep network model \cite{xiao2016end},
where results were obtained by running the code provided by the authors. Our
obtained results closely match those reported in \cite{xiao2016end} (Figure 7
b)). Overall, our DLDP model consistently ourperforms the end-to-end person
search deep model by a constant 2\% in mAP regardless gallery size; 3\% in
Rank-1 for low gallery sizes of 50 images (correspoding to 256 bounding boxes)
and up to 5.4\% in Rank-1 for the largest possible gallery of all 6978 images
(36984 bounding boxes). This suggests that the DLDP model is
less sensitive to increase in gallery size, even {\em without}
benefiting from learning on target domains.

\noindent {\bf Effects from occlusion and low-resolution}:
Finally, we evaluated the effects of occlusion and low-resolution probe
images. The CUHK-SYSU dataset provides two probe subsets for this purpose,
which were created by sampling heavily occluded probe images and the 10\% probe
images with the lowest resolutions, respectively. Gallery sizes for this
evaluation are fixed at 100 images. We report results using the SSD VOC500
person detection in Table \ref{tab:eval_occres} and compared to the end-to-end
person search deep network model. Consistent with the
observation made by \cite{xiao2016end}, an occluded probe image
causes more difficulty for re-id than that of low-resolution imagery. For
low-resolution, our DLDP model suffers only a 15\% loss in mAP and Rank-1, as
compared to a 20\% decrease for the end-to-end person search deep model. On
occlusion, the reported results on the end-to-end person search model are less
affected (reduced by 16.0\% mAP and 19.4\% Rank-1) than our DLDP model whose
performance is reduced by 18.8\% in mAP and 25.6\% in Rank-1.
However, using the author provided code, we could not re-create the same
results as reported in \cite{xiao2016end} for the occlusion test. Instead, we
obtained the result on the end-to-end person search model for occlusion test
5\% lower than reported, almost identical to that of DLDP. 

\begin{table*}[ht!]
\begin{center}
\begin{tabular}{|l|c|c|c|c|c|}
\hline
                    & mAP   & Rank-1 & Rank-5 & Rank-10 & Rank-20 \\
\hline\hline
DLDP GT Whole       & 74.0  & 76.7   & 86.4   & 89.7    & 92.9    \\
DLDP GT Occlusion   & 56.0  & 54.0   & 69.5   & 76.5    & 83.4    \\
DLDP GT LowRes      & 72.0  & 74.1   & 86.9   & 89.7    & 93.1    \\
\hline
\end{tabular}
\vspace{.1cm}
\caption{Results of DLDP on the occlusion and low resolution test sets
         using ground-truth detections.}
\label{tab:eval_occres_gt}
\end{center}
\end{table*}

To gain more insight, we further report the re-id results from our DLDP model
on the occlusion and low-resolution tests but this time using manually labelled
person bounding boxes (Table \ref{tab:eval_occres_gt}). The overall results are
much improved by relying on ground truth detections. This may partly be due to
that ground truth labelled bboxes resemble more closely to data the model was
trained on, therefore the negative impact of low-resolution query images is
less severe. This suggests that the resolution gap between probe and gallery
can be handled well by DLDP provided person bbox detection is reasonably
accurate without significant misalignment. 

\begin{figure*}[ht!]
    \begin{center}
        \includegraphics[width=0.99\linewidth]{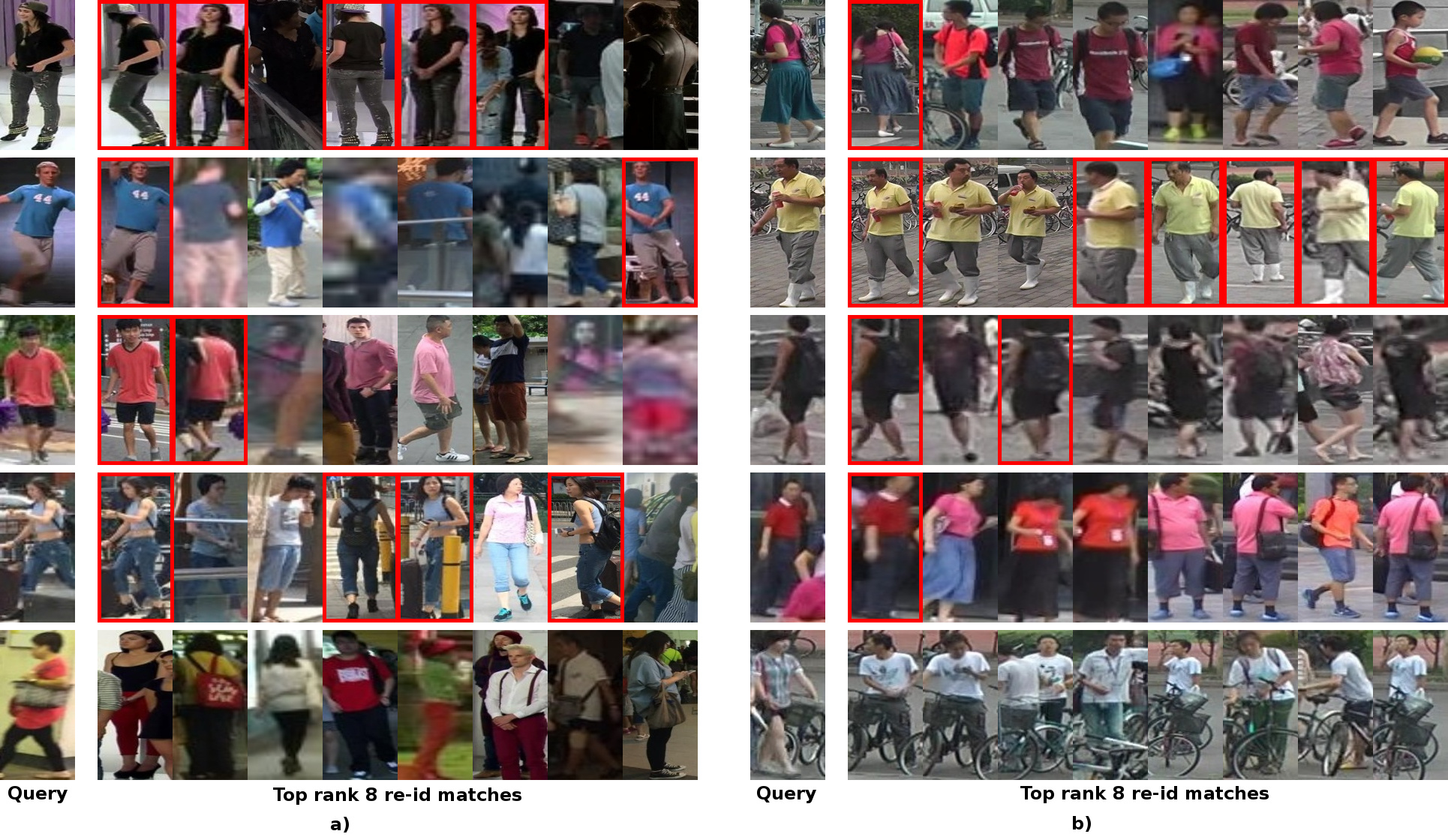}
    \end{center}
    \caption{(a) The top 8 re-id matches by the DLDP model on the CUHK-SYSU test 
             data for a set of five randomly chosen queries from the 100 image 
             gallery setting, and (b) five randomly chosen queries on the PRW test 
             data. Note, rank-2 and rank-3 in the ``yellow T-shirt'' example in (b) are 
             false matches even though they look very similar. The bottom examples 
             from both (a) and (b) show failure cases when the model failed to find a 
             match in the top 8 ranks.}
    \label{fig:qualitative}
\end{figure*}

\section{Conclusion}

In this work, we presented a novel approach to domain sensitive person
re-identification by deep learning {\em without} the need for training
data from the target (test) domains. The new model DLDP automatically
discovers prototype domains from independent diverse datasets and
learns specific feature embedding for each of the discovered
domains. In model deployment, each query image is used to select for
the most suitable feature embedding with its corresponding domain best
fitting the query before the ranking match against the gallery
candidates. Our approach has a singificant {\em unique} advantage, over
{\em all} existing models, of not requiring any target domain data for
model learning.
Our extensive comparative evaluation on two latest benchmark datasets 
demonstrate clearly that the proposed DLDP model
outperforms the state-of-the-art or is competitive, notwithstanding
that all other models benefit from having been trained on the
target domain data.
It is also evident that the proposed new DLDP model copes well with
real-world re-id conditions when automatic person detection,
occlusion, low resolution and very large gallery sizes (\ie open
world) are unavoidable in model deployment.
Future work includes investigating in more detail the impact of the number
of domains on the accuracy of our approach as well as ways of coupling the
domain discovery and learning of domain embeddings more directly in an
end-to-end approach.

\ifwacvfinal
\section{Acknowledgements}
This work was supported in part by the research travel grant of the
Karlsruhe House of Young Scientists (KHYS). 
\fi

{\small
\bibliographystyle{ieee}
\bibliography{egbibarxiv}
}

\newpage

\section*{Appendix}

\subsection*{Evaluation on Market-1501}

We additionally evaluted DLDP on the Market-1501 \citeadx{zheng2015scalable} dataset. 
This dataset does not
provide full images but was created using the DPM detector instead of manual
annotations. Results of DLDP in comparison to state-of-the-art approaches are given
in Table \ref{tab:eval_sota_market}.
DLDP outperforms many recent approaches and performs en-par with the approach
of \citeadx{chen2016similarity} (-0.12\% mAP and -0.44 Rank-1). DLPD is clearly
outperformed by \citeadx{Zhang_2016_CVPR} and \citeadx{varior2016gated} which both
achieve a significant improvement over the previous state-of-the-art and beat
DLPD by up to 13.32\% in mAP and 14.42\% in Rank-1. All approaches given in
Table \ref{tab:eval_sota_market} make full use of the Market training dataset
with the noteable exception of \citeadx{martinel2016temporal} which uses only
13.58\% of the training data to adapt model pretrained on other data. DLPD
clearly outperforms this result without using any of the training data.

A qualitative impression of the results of DLPD on the Market-1501 dataset is
depicted in Figure \ref{fig:qualitative}. Again,
it can be observed that most of the incorrect results are quite reasonable and
share one or more salient features with the query person (e.g. the black
backpack in the second row or either a bag-strap or gray shirt in the fourth
row). The last row shows a query for which only one out of five true matches
is returned among the top 15 results. We consider this a near failure case.
A success implies that the model finds \textit{all} of the existing true 
matches.

\begin{table}[ht!]
\begin{center}
\begin{tabular}{|l|c|c|}
\cline{2-3}
\multicolumn{1}{c|}{}
                                                            & mAP       & Rank-1   \\
\cline{2-3}
\noalign{\vskip .08cm}
\hline
Gated S-CNN \citeadx{varior2016gated}                          & \f{39.55} & \f{65.88}\\
DNS \citeadx{Zhang_2016_CVPR}                                  & 35.68     & 61.02    \\
SCSP \citeadx{chen2016similarity}                              & 26.35     & 51.90    \\
\hline
DLDP                                                        & 26.23     & 51.46    \\    
\hline
Multiregion Bilinear DML \citeadx{ustinova2015multiregion}     & 26.11     & 45.58    \\
End-to-end CAN \citeadx{liu2016end}                            & 24.43     & 48.24    \\
TMA LOMO \citeadx{martinel2016temporal}                        & 22.31     & 47.92    \\
WARCA-L \citeadx{jose2016scalable}                             & -         & 45.16    \\
MST-CNN \citeadx{liu2016multi}                                 & -         & 45.1     \\
\hline
\end{tabular}
\vspace{.1cm}
\caption{DLDP's performance in context of many recent state-of-the-art approaches for the 
         single-query setting on the Market-1501 dataset.} 
\label{tab:eval_sota_market}
\end{center}
\end{table}

\subsection*{Ground-Truth Detections on PRW}

In Table \ref{tab:eval_gt_prw} we show the performance of DLDP and our baseline
model (see Section 3.2.1 in the main paper) on the PRW dataset. We compare to 
the BoW+XQDA baseline which is provided
with the evaluation code of \citeadx{zheng2016person}. Compared to the CUHK-SYSU
dataset (compare Table 3, main paper) the improvement in accuracy achieved by 
relying on ground-truth is much less pronounced for all approaches. This is 
likely due to the fact that the ground-truth on PRW was obtained in part using 
the DPM-Inria person detector, thus giving that detector an unusually high 
localization accuracy on the dataset. This also explaines the comparatively weak
performance of the SSD detectors observed in Table 4 of the main paper.
DLDP is able to maintain its advantage over the BoW+XQDA baseline on ground-truth
and outperforms it by 1.8\% mAP and 8.5\% Rank-1.

\begin{table}[ht!]
\begin{center}
\begin{tabular}{|l|c|c|}
\cline{2-3}
\multicolumn{1}{c|}{}
                                        & mAP       & Rank-1    \\
\cline{2-3}
\noalign{\vskip .08cm}
\hline
BoW + XQDA                              &    16.7   &    38.8  \\
Baseline Model                          &    16.1   &    43.2  \\
DLDP                                    & \f{18.5}  & \f{47.3} \\
\hline
\end{tabular}
\vspace{.1cm}
\caption{DLDP re-id performance on the PRW dataset using ground-truth annotations 
         instead of automatic detections.}
\label{tab:eval_gt_prw}
\end{center}
\end{table}


\subsection*{Full CMC-Curves}

\begin{figure}[ht!]
    \begin{center}
        \includegraphics[width=0.49\textwidth]{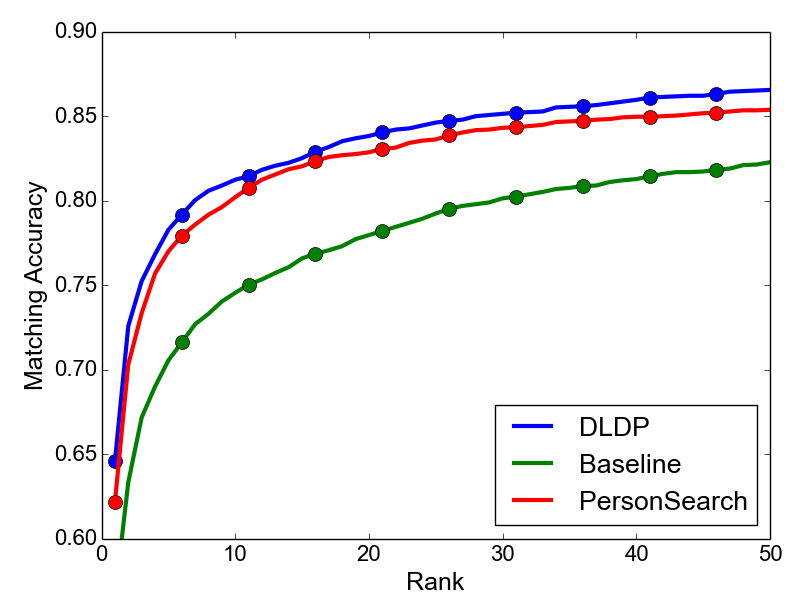}
    \end{center}
    \caption[asd]{CMC curve of DLDP on CUHK-SYSU compared to our baseline model
             and the PersonSearch approach presented in \citeadx{xiao2016end}.
             Results are obtained using the default gallery size of 100 images
             and the SSD-VOD500 detector for DLDP and the baseline model.}
    \label{fig:cmc_sysu}
\end{figure}

\begin{figure}[ht!]
    \begin{center}
        \includegraphics[width=0.49\textwidth]{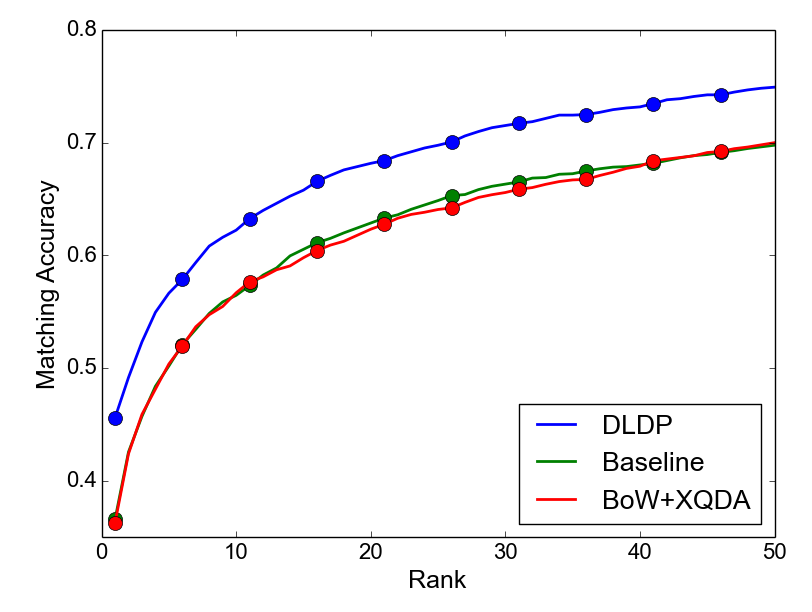}
    \end{center}
    \caption{CMC curve of DLDP on PRW compared to our baseline model and the
             BoW+XQDA baseline provided with the evaluation code. Results were
             obtained by considering the top 5 detections in each image. All 
             approaches are evaluated on the provided detections of the DPM-Inria 
             detector.}
    \label{fig:cmc_prw}
\end{figure}

\begin{figure}[ht!]
    \begin{center}
        \includegraphics[width=0.49\textwidth]{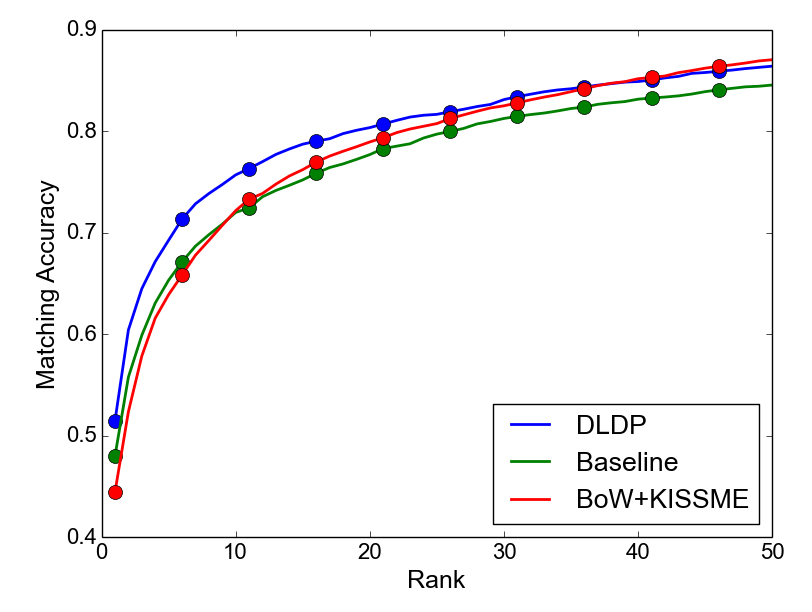}
    \end{center}
    \caption{CMC curve of DLDP on Market-1501 compared to our baseline model and
             the BoW+KISSME baseline provided with the evaluation code. Results
             are for the single-query setting.}
    \label{fig:cmc_market}
\end{figure}

In Figures \ref{fig:cmc_sysu}, \ref{fig:cmc_prw} and \ref{fig:cmc_market} 
we give the full CMC curves for our main experiments on CUHK-SYSU, PRW, 
and Market-1501, respectively. All curves were 
generated using the evaluation code provided with the datasets and are 
compared to the strongest baseline approaches for which code was provided.

In Figure \ref{fig:cmc_sysu} we show DLDP's average accuracy over the first 
50 ranks on the CUHK-SYSU dataset. DLDP shows a consistent improvement 
of more than 5\% over
the baseline model and narrowly but consistently outperforms the deep 
PersonSearch approach which integrates detection and re-id into one CNN.

Figure \ref{fig:cmc_prw} shows the first 50 ranks on the PRW dataset. Our 
baseline model performs en-par with the BoW+XQDA baseline. Both appraoches
are again consistently outperformed by more than 5\% by DLDP.

The average accuracy over the first 50 ranks on the Market-1501 dataset is 
depicted in Figure \ref{fig:cmc_market}. The difference between DLDP, our
baseline and the BoW+KISSME baseline is less significant on this dataset.
However, in the important segment of ranks 1-20 DLDP has a clear advantage.
For ranks above 38 the BoW+KISSME approach actually outperforms both our 
baseline model and DLDP narrowly.

\begin{figure*}[ht!]
    \begin{center}
        \includegraphics[width=0.99\linewidth]{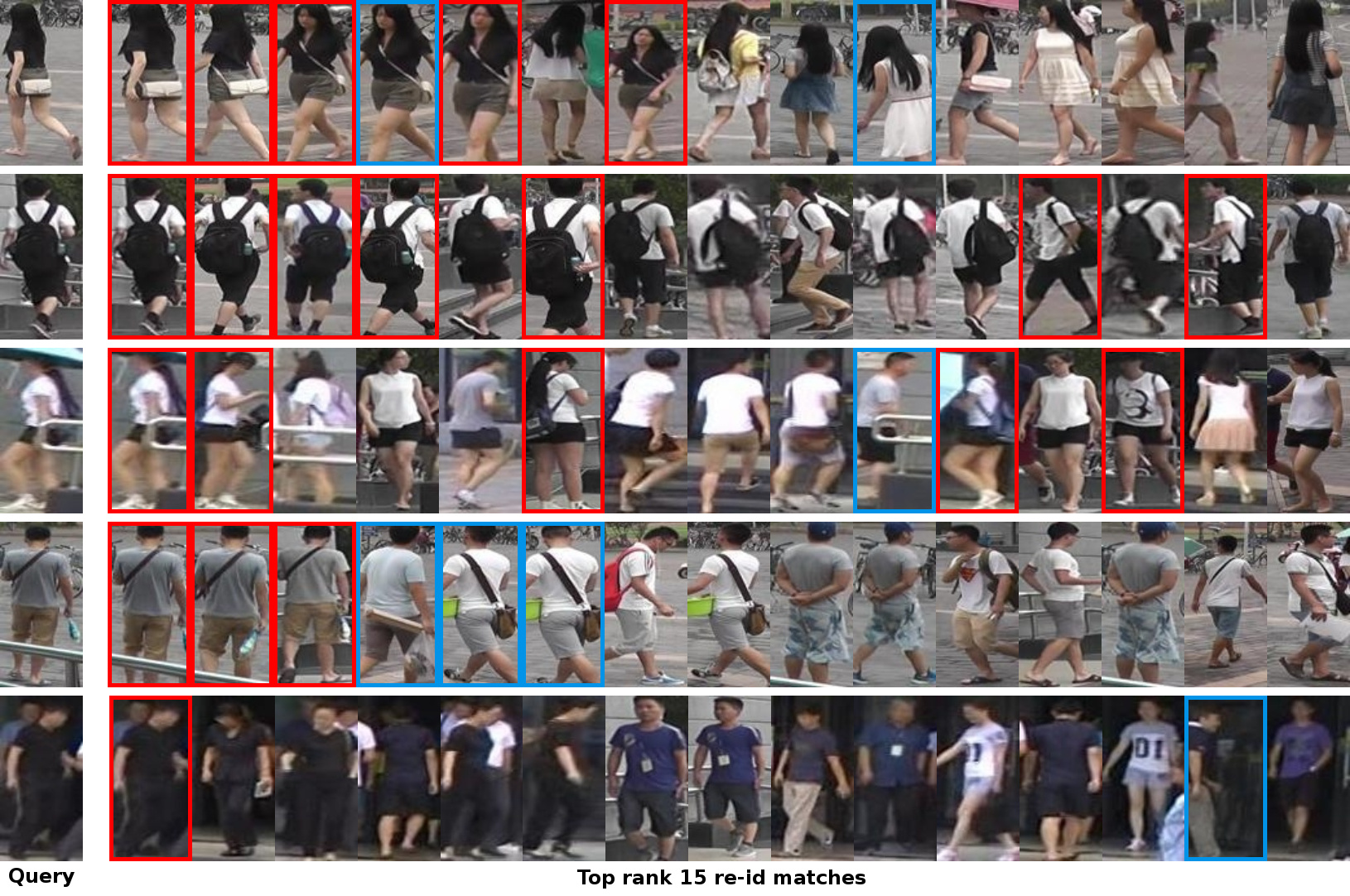}
    \end{center}
    \caption{The top 15 re-id matches by the DLDP model on the Market-1501 dataset. Correct matches 
             are framed red. Matches labelled as ``junk'' in the dataset and not considered in the 
             evaluation protocol are framed blue. The last row shows a failure case where only one
             (out of five) correct images could be found in the top 15 results.}
    \label{fig:qualitative}
\end{figure*}

{\small
\bibliographystyleadx{ieee}
\bibliographyadx{supplementary/egbib}
}

\end{document}